\documentclass{article}

\usepackage[preprint]{neurips_2025}

\usepackage[utf8]{inputenc} 
\usepackage[T1]{fontenc}    
\usepackage[colorlinks,linkcolor=black,anchorcolor=blue,citecolor=green]{hyperref}        
\usepackage{url}            
\usepackage{booktabs}       
\usepackage{amsfonts}       
\usepackage{nicefrac}       
\usepackage{microtype}      
\usepackage{enumitem}
\usepackage{amsmath}
\usepackage{bm}
\usepackage{graphicx}
\usepackage{subcaption}

\usepackage{multirow}
\usepackage{graphicx}
\usepackage{caption}
\usepackage{bbm}
\usepackage{multirow}
\usepackage{wrapfig}
\usepackage{colortbl}
\usepackage{color}
\usepackage[table,xcdraw]{xcolor}
\usepackage[normalem]{ulem}
\useunder{\uline}{\ul}{}
\def\eg{\emph{e.g.}}
\def\ie{\emph{i.e.}}

\newcommand{\etal}{\textit{et} \textit{al}.\space}

\newcommand{\tool}{\textsc{ArgRE}\space}
\newcommand{\toolns}{\textsc{ArgRE}}

\usepackage{cleveref}
\Crefname{figure}{Fig.}{Figs.} 
\crefname{figure}{Fig.}{Figs.} 
\crefname{table}{Tab.}{Tabs.}
\Crefname{table}{Tab.}{Tabs.}

\usepackage{pifont}
\usepackage[perpage,symbol*]{footmisc}
\DefineFNsymbols{circled}{{\ding{192}}{\ding{193}}{\ding{194}}
{\ding{195}}{\ding{196}}{\ding{197}}{\ding{198}}{\ding{199}}{\ding{200}}{\ding{201}}}

\newcommand\myfootnotestyle[1]{\ifcase#1 \or \ding{182}\or \ding{183}\or
\ding{184}\or \ding{185}\or \ding{186}\or \ding{187}%
\or \ding{188}\or \ding{189}\or \ding{190}\or \ding{191}\else *\fi\relax}

\newcommand*{\affaddr}[1]{#1} 
\newcommand*{\affmark}[1][*]{\textsuperscript{#1}}

\title{Detoxifying Large Language Models via Autoregressive Reward Guided Representation Editing}

\author{
    Yisong Xiao\affmark[1],
    \textbf{Aishan Liu\affmark[1]\thanks{Corresponding Author},}  
    Siyuan Liang\affmark[2], 
    \textbf{Zonghao Ying\affmark[1]}, 
    \textbf{Xianglong Liu\affmark[1,3,4]},
    \textbf{Dacheng Tao\affmark[5]}
    \\
    \affaddr{\affmark[1]SKLCCSE, Beihang University}
    \affaddr{\affmark[2]{National University of Singapore}} \\
    \affaddr{\affmark[3]{Zhongguancun Laboratory, Beijing}}
    \affaddr{\affmark[4]{Institute of Dataspace, Hefei}}
    \affaddr{\affmark[5]{Nanyang Technological University}}
}

\begin{document}

\maketitle

\begin{abstract}
Large Language Models (LLMs) have demonstrated impressive performance across various tasks, yet they remain vulnerable to generating toxic content, necessitating detoxification strategies to ensure safe and responsible deployment. Test-time detoxification methods, which typically introduce static or dynamic interventions into LLM representations, offer a promising solution due to their flexibility and minimal invasiveness. However, current approaches often suffer from imprecise interventions, primarily due to their insufficient exploration of the transition space between toxic and non-toxic outputs. To address this challenge, we propose \textsc{A}utoregressive \textsc{R}eward \textsc{G}uided \textsc{R}epresentation \textsc{E}diting (\toolns), a novel test-time detoxification framework that explicitly models toxicity transitions within the latent representation space, enabling stable and precise reward-guided editing. \tool identifies non-toxic semantic directions and interpolates between toxic and non-toxic representations to reveal fine-grained transition trajectories. These trajectories transform sparse toxicity annotations into dense training signals, enabling the construction of an autoregressive reward model that delivers stable and precise editing guidance. At inference, the reward model guides an adaptive two-step editing process to obtain detoxified representations: it first performs directional steering based on expected reward gaps to shift representations toward non-toxic regions, followed by lightweight gradient-based refinements. Extensive experiments across 8 widely used LLMs show that \tool significantly outperforms leading baselines in effectiveness (-62.21\% toxicity) and efficiency (-47.58\% inference time), while preserving the core capabilities of the original model with minimal degradation. Our code is available at the \href{https://anonymous.4open.science/r/ARGRE-6291}{website}.


\end{abstract}

\vspace{-0.05in}
\section{Introduction}
\vspace{-0.05in}

Large Language Models (LLMs) have made substantial progress, showcasing remarkable capabilities across various domains and tasks \cite{brown2020language,chiang2023vicuna,touvron2023llama2,achiam2023gpt}. 
Despite these achievements, LLMs continue to face significant challenges related to toxicity \cite{ying2025jailbreak,ying2025pushing,ying2024safebench,ying2025reasoning,zou2025prism}, robustness \cite{liu2020spatiotemporal,liu2023x,zhang2021interpreting,xiao2023robustmq}, and other trustworthiness concerns \cite{xiao2025bdefects4nn,ying2023dlp,liu2025agentsafe,xiao2023latent,liu2025elba, wang2025manipulating}. This paper specifically addresses the notorious toxicity issues associated with LLMs, which remain vulnerable to generating \emph{harmful or toxic content}, primarily due to their pre-training on large, unfiltered text corpora that may inadvertently encode harmful patterns \cite{deshpande2023toxicity,gehman2020realtoxicityprompts, liang2025revisiting,liang2025vl,liu2025natural}. As LLMs are increasingly integrated into socially sensitive applications, developing effective detoxification techniques is critical to ensuring their ethical and responsible deployment \cite{liang2025safemobile,liang2024unlearning}.

A significant body of research has focused on mitigating toxicity in LLMs \cite{gururangan2020don,zhang2023mil,lidestein,uppaalmodel,lee2024mechanistic, lu2025adversarial, liang2025t2vshield}. Prior studies \cite{wang2022exploring,gururangan2020don,wang2024secrets} involve fine-tuning LLMs on carefully curated preference datasets (pairs of toxic and non-toxic responses) using algorithms like direct preference optimization (DPO) \cite{rafailov2023direct}. However, these training-time methods require costly data collection and substantial computational resources, making them impractical in low-resource scenarios. Consequently, recent work has shifted towards \emph{test-time detoxification} during inference, with representation editing \cite{lidestein,lee2024mechanistic,leong2023self,kong2024aligning,xiao2025fairness,xiao2025genderbias} gaining widespread attention for its flexibility and minimal invasiveness. Building upon the linear representation hypothesis \cite{park2024geometry,park2024linear,nanda2023emergent}, which posits that human-interpretable concepts are encoded as linear directions within LLM representations, representation editing methods guide representations through static or dynamic interventions toward non-toxic directions to suppress toxic behaviors. However, these methods are often limited by imprecise interventions, primarily due to insufficient exploration of the transition space between toxic and non-toxic outputs. In particular, reliance on sparse toxicity annotations prevents these methods from capturing the nuanced intermediate transitions necessary for stable and precise guidance.

To address this challenge, we propose \textbf{\textsc{A}}utoregressive \textbf{\textsc{R}}eward \textbf{\textsc{G}}uided \textbf{\textsc{R}}epresentation \textbf{\textsc{E}}diting (\toolns), a test-time detoxification framework that explicitly models toxicity transitions within the latent representation space, enabling stable and precise reward-guided editing. Leveraging the continuous semantic representation space, we can track and characterize toxicity shifts, which allows for the exploration of toxicity transition trajectories that are difficult to capture in the discrete natural language space. Specifically, \tool first identifies the non-toxic semantic direction and then interpolates between toxic and non-toxic representations along this direction to uncover fine-grained toxicity transition trajectories. 
These trajectories convert sparse toxicity annotations into dense pairwise training signals, facilitating smooth transitions across toxicity levels. Leveraging these trajectories, we develop an autoregressive reward model that estimates the toxicity of token representations, providing stable and precise guidance for editing. During generation, \tool employs an adaptive two-step editing process: it first steers the representation toward non-toxic regions based on the expected reward gap, followed by lightweight gradient ascent to further maximize the reward (\ie, reduce toxicity), achieving effective and efficient detoxification.

Extensive experiments across eight widely-used LLMs show that \tool consistently delivers strong detoxification, reducing toxicity by up to 62.21\%, while demonstrating great efficiency by decreasing inference overhead by 47.58\% compared to the leading test-time methods. In addition, \tool preserves the original capabilities of the LLM with minimal impact on overall performance. Benefiting from toxicity transition exploration, \tool also exhibits high data efficiency, without requiring extensive data annotations. Beyond detoxification, we explore its applicability to stereotype recognition and jailbreak mitigation, observing promising results.
Our main \textbf{contributions} are:
\begin{itemize}[leftmargin=2em]
    \item We propose \toolns, a test-time detoxification framework that models toxicity transitions within the latent representation space to enable stable and precise representation editing guidance.
    \item We develop an autoregressive reward model to evaluate the toxicity of token representations and design an adaptive two-step editing strategy for effective and efficient detoxification.
    \item Extensive experiments demonstrate that \tool significantly outperforms leading baselines in both effectiveness and efficiency, while maintaining LLM's core capabilities.
\end{itemize}

\vspace{-0.1in}
\section{Related Works}
\vspace{-0.075in}


\textbf{Training-time methods} mitigate toxicity by modifying LLM parameters through fine-tuning on curated non-toxic datasets \cite{wang2022exploring,gururangan2020don,liu2023towards,liu2023exploring,liu2021training}. For instance, Wang \etal \cite{wang2024secrets} apply reinforcement learning from human feedback (RLHF) \cite{ouyang2022training} to calibrate harmful generation based on preference data. DPO \cite{rafailov2023direct} streamlines this by directly fine-tuning on preference pairs. However, these methods rely on large-scale data and intensive computation, limiting their democratization and broader applicability.

\textbf{Test-time methods} generally fall into three categories. 
\ding{182} \textit{Guided decoding methods} \cite{dathathri2019plug,liu2021dexperts,krause2020gedi,zhang2023mil,xu2024genarm} modify the token probability distribution in frozen LLMs during decoding to mitigate toxic generations. DexPerts \cite{liu2021dexperts} employs trained classifiers to distinguish between toxic and non-toxic attributes, promoting the selection of tokens aligned with non-toxic traits. GenARM \cite{xu2024genarm} eliminates the need for external classifiers by learning a reward model that aligns with the base LLM to score token probability distributions, enabling efficient generation toward more desirable outcomes. However, directly altering token probabilities can disrupt natural generation, leading to degraded fluency and coherence under strong control.  
\ding{183} \emph{Weight editing methods} \cite{wei2024assessing,uppaalmodel,wang2024detoxifying} detoxify LLMs by removing harmful components from their parameters, such as ProFS \cite{uppaalmodel}, which uses low-rank decomposition and projection to isolate and eliminate toxic MLP weights. However, weight editing may lead to degraded detoxification performance on large-scale language models and risk compromising their general capabilities \cite{gu2024model}.
\ding{184} \emph{Representation editing methods} \cite{leong2023self,panickssery2023steering,liu2023context,lidestein,lee2024mechanistic,kong2024aligning} mitigate toxicity by applying targeted interventions to LLM representations. Self-Detoxify \cite{leong2023self} identifies toxic directions by contrasting toxic and non-toxic examples, applying static interventions during inference to suppress toxicity. DeStein \cite{lidestein} enhances this by using linear classifiers trained on a few toxicity annotations for more precise toxic direction identification. Re-Control \cite{kong2024aligning} improves static edits by learning a value function that generates dynamic intervention signals, guiding tedious gradient-based iterations to achieve desirable representations. However, these methods often suffer from imprecise interventions, as they fail to adequately explore the transition space between sparse toxicity annotations, leading to suboptimal performance.

Our \tool \textbf{distinguishes} itself in three key aspects:
\ding{182} \textit{Motivation}. \tool explicitly models toxicity transitions within the representation space, constructing dense trajectories that enable more effective detoxification, whereas prior methods depend on sparse toxicity annotations with insufficient transition exploration.
\ding{183} \textit{Implementation}. Leveraging these transitions, \tool learns precise and stable rewards to guide an adaptive two-step representation editing process, avoiding the imprecise interventions and intrusive token- or weight-level modifications of existing approaches.
\ding{184} \textit{Effects}. \tool consistently achieves strong performance with high efficiency, while existing methods are often constrained by suboptimal effectiveness or substantial computational overhead.

\vspace{-0.075in}
\section{Methodology}
\begin{figure*}[t]
\centering
\includegraphics[width=0.95\linewidth]{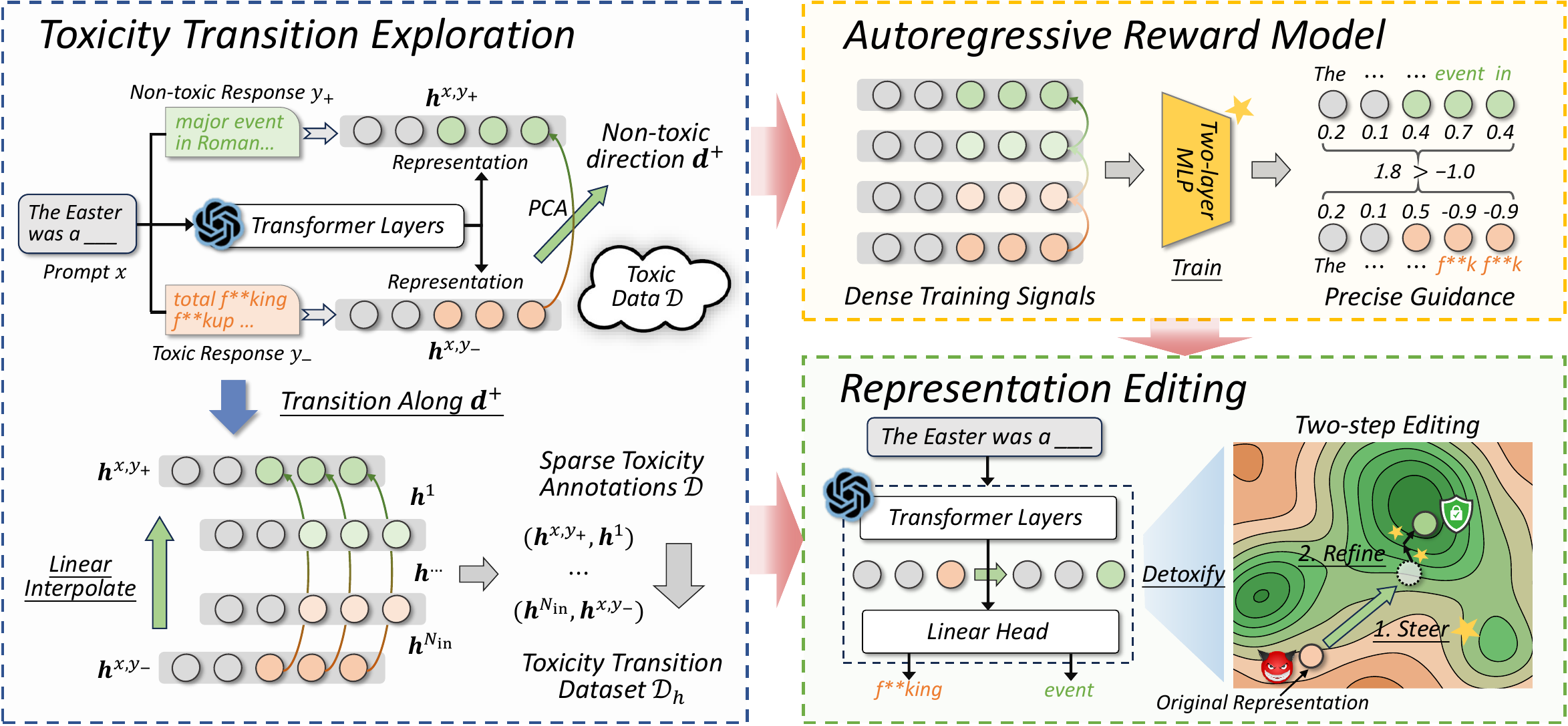}
\caption{Overview of \toolns. It identifies non-toxic semantic directions and interpolates between toxic and non-toxic representations to reveal fine-grained transition trajectories. These trajectories transform sparse toxicity annotations into dense training signals, enabling the construction of an autoregressive reward model that delivers stable and precise editing guidance. At inference, the reward model guides an adaptive two-step editing process to obtain detoxified representations.}
\label{fig:framework}
\vspace{-0.15in}
\end{figure*}

\vspace{-0.05in}
In this section, we first briefly review RLHF fundamentals to understand non-toxicity editing; then, we present our \toolns, which explicitly models toxicity transitions in the representation space, transforming sparse toxicity annotations into dense training signals, thereby facilitating the learning of an autoregressive reward model that provides stable and precise guidance for editing. An overview of the framework is provided in \cref{fig:framework}.

\vspace{-0.05in}
\subsection{Preliminaries and Motivation}
\vspace{-0.025in}

\textbf{Reward model learning based on pairwise toxic data.} Typically, a reward model $r(x, y)$ outputs a scalar score given a prompt $x = \{x_1, \dots, x_M\}$ and response $y = \{y_1, \dots, y_T\}$, where $x_{m}$ and $y_{t}$ denote the tokens of $x$ and $y$, respectively. Given a pairwise toxic dataset $\mathcal{D}$ consisting of triples $(x, y_{+}, y_{-})$, where $y_{+}$ and $y_{-}$ denote the non-toxic and toxic responses generated by the base LLM $\pi_{\rm{base}}(y \mid x)$, the reward model is trained by minimizing the negative log-likelihood loss to encourage higher scores for non-toxic responses:
\begin{equation}
\label{eq:reward_model_loss}
    \min_{r} -\mathbb{E}_{(x, y_{+}, y_{-})\sim \mathcal{D}}\bigl[\log \sigma(r(x, y_{+})- r(x, y_{-}))\bigr], 
\end{equation}
where $\sigma$ denotes the logistic function. The reward model $r(x,y)$ is usually initialized from the base LLM $\pi_{\rm{base}}(y \mid x)$, with a trainable linear layer $\theta_{l}$ stacked on top of the final transformer layer \cite{ziegler2019fine}.

\textbf{Kullback-Leibler (KL)-regularized RL fine-tuning.} Using the reward model, RLHF fine-tunes the base LLM $\pi_{\rm{base}}(y \mid x)$ to mitigate toxicity by maximizing expected reward while minimizing the KL divergence from the base model: 
\begin{equation}
\label{eq:RL_objective}
\max_\pi \mathbb{E}_{x\sim\mathcal{D}, y\sim \pi(x)} r(x,y) - \beta D_{\rm{KL}}(\pi(y|x)||\pi_{\rm{base}}(y|x)),
\end{equation}
where $\beta$ is a hyperparameter that controls the trade-off between reward maximization and preserving the behavior of the base LLM. 
Following prior work \cite{go2023aligning,rafailov2023direct}, the objective in Eqn~\ref{eq:RL_objective} admits a closed-form solution, given by: 
\begin{equation}
\label{eq:closed_form}
 \hat{\pi}(y|x) \propto \pi_{\text{base}}(y|x) \exp \bigl(\frac{1}{\beta} r(x,y)\bigl),    
\end{equation}
where $\pi_{\text{base}}$ remains frozen, $y$ denotes any potential response, and the reward $r(x,y)$ guides the base LLM's generation to produce a modulated distribution $\hat{\pi}$ that favors high-reward (\ie, non-toxic) responses. Specifically, the reward is computed from the base LLM's final-layer hidden representation $\bm{h}^{x,y}$ as: $r(x,y)=\theta_{l}(\bm{h}^{x,y})$. Thus, the representation $\bm{h}^{x, y}$ directly influences the reward score and,  consequently, the toxicity of the generated content, serving as a minimally invasive interface for controlling the LLM's output toward non-toxic responses.

\textbf{Motivation.} Existing representation editing methods \cite{lidestein,leong2023self,kong2024aligning,panickssery2023steering} steer the representation $\bm{h}^{x, y}$ toward non-toxic regions (\ie, high-reward areas) via static or dynamic interventions. However, due to limited exploration of transitions between toxic and non-toxic outputs, such interventions are often imprecise, leading to suboptimal reward scores and detoxification performance. To address this, we explicitly model toxicity transitions within the latent representation space, transforming sparse toxicity annotations into dense training signals to enable stable and precise reward-guided editing.



\vspace{-0.025in}
\subsection{Toxicity Transition Exploration}
\vspace{-0.025in}

Building on the linear representation hypothesis \cite{park2024geometry,park2024linear,nanda2023emergent}, which suggests that concepts like toxicity are encoded as linear directions in LLM representations, we efficiently capture toxicity transitions by exploring the continuous semantic representation space. Specifically, we first identify the non-toxic direction and then interpolate along it to trace how toxicity evolves, bridging sparse annotations to uncover fine-grained transition trajectories.

Given a prompt $x$ and its corresponding response $y$, the final-layer representation of the LLM, $\bm{h}^{x,y}$, can be decomposed as $\bm{h}^{x,y} = \{\bm{h}_{[1]},\dots,\bm{h}_{[M]},\bm{h}_{[M+1]},\dots,\bm{h}_{[M+T]}\}$. Therefore, for a prompt $x$ with non-toxic response $y_{+}$ and toxic response $y_{-}$, the non-toxic direction can be derived from their representation difference at the last token:
\begin{equation}
    \Delta\bm{h}{(x,y_{+},y_{-})} = \bm{h}_{[-1]}^{x,y_{+}}-\bm{h}_{[-1]}^{x,y_{-}}.
\end{equation} 
We only utilize the last token representation to determine direction, as the LLM is causally modeled and the attention mechanism aggregates information from all tokens into the last one \cite{meng2022locating}. To enhance generalizability across different toxic pairs, we aggregate the non-toxic direction matrix $\{\Delta\bm{h}({x^{(i)},y_{+}^{(i)},y_{-}^{(i)}})\}_{i=1}^{N}$ from a small sample set (size $N$), and apply PCA \cite{shlens2014tutorial} to identify the first principal component $\bm{d}_{+}$, which captures the dominant non-toxic direction.

The direction $\bm{d}_{+}$ offers a clear path for exploring the transitions between non-toxic and toxic pairs ($\bm{h}^{x,y_{+}}$ and $\bm{h}^{x,y_{-}}$) in the high-dimensional semantic representation space. Specifically, we perform linear interpolation at the token level between $\bm{h}^{x,y_{+}}$ and $\bm{h}^{x,y_{-}}$ along the non-toxic direction:
\begin{equation}
\label{eq:interpolation}
    \begin{aligned}
    \bm{h}^{\lambda}_{[t]} = 
    \begin{cases}
    \bm{h}_{[t]}^{x,y_{+}}, & t \in [1,\dots,M]\\
    \bm{h}_{[t]}^{x,y_{+}} + \frac{\lambda}{N_{\rm{in}}+1} \cdot [\bm{d}^{\rm{T}}_{+} (\bm{h}_{[t]}^{x,y_{-}} - \bm{h}_{[t]}^{x,y_{+}})] \cdot \bm{d}_{+}, & t \in [M+1,\dots,M+T]
    \end{cases}
    \end{aligned}
\end{equation}
where $N_{\rm{in}}$ is the number of interpolated trajectories, $\lambda \in [1,\dots,N_{\rm{in}}]$, and $\bm{h}^{\lambda}$ is one of the interpolated toxicity transition trajectories $\{\bm{h}^{\lambda}\}_{\lambda=1}^{N_{\rm{in}}}$. For token positions $t \in [1,\dots,M]$, the representation remains unchanged as the input is solely related to the prompt $x$. For $t \in [M+1,\dots, M+T]$, we first project the representation difference between $\bm{h}^{x,y_{+}}$ and $\bm{h}^{x,y_{-}}$ onto the non-toxic direction, and then interpolate along this direction to generate transition trajectories. In practice, interpolation stops when the shorter token sequence between $y_{+}$ and $y_{-}$ is reached.

These trajectories serve as dense supervision signals, transforming sparse toxicity annotations into fine-grained transitions from toxic to non-toxic representations. Based on them, we construct a pairwise representation-level dataset $\mathcal{D}_{h}$:
\begin{equation}
\mathcal{D}_{h}=\bigcup_{(x, y_{+}, y_{-}) \in \mathcal{D}} \left\{(\bm{h}^{x,y_{+}},\bm{h}^{1}), (\bm{h}^{1},\bm{h}^{2}), \dots, (\bm{h}^{N_{\rm{in}}},\bm{h}^{x,y_{-}})\right\}.
\end{equation}
Compared to the original dataset $\mathcal{D}$, our constructed dataset $\mathcal{D}_{h}$ captures denser toxicity transitions, enabling the learning of a reward model that provides more stable and accurate guidance.


\vspace{-0.025in}
\subsection{Autoregressive Reward Model Construction}
\vspace{-0.025in}


Trajectory-level reward models are trained on complete trajectories and assign the final reward only at the last token, resulting in imprecise editing signals during generation \cite{sutton1998reinforcement,pignatellisurvey}. In contrast, we train an autoregressive reward model that operates at the token level, providing more fine-grained and precise guidance for representation editing. Specifically, our autoregressive reward model $\theta_{r}$ assigns a scalar reward to each token representation, decomposing the overall reward $r(x, y)$ into a sum over token-wise representation rewards:
\begin{equation}
\label{eq:reward_decompose}
    r(x,y)=\sum_{t=1} \theta_{r}(\bm{h}^{x,y_{\le t}}_{[M+t]}), 
\end{equation}
where $\bm{h}^{x,y_{\leq t}}_{[M+t]}$ is the representation of the $t$-th token, which implicitly depends on all preceding representations $\bm{h}^{x,y_{\leq t}}_{[<M+t]}$ due to the auto-regressive nature of the base LLM. 

Our autoregressive reward model $\theta_{r}$ is implemented as a learnable two-layer MLP applied on top of the final transformer layer. It is trained on the dense toxicity transition dataset $\mathcal{D}_{h}$ using an objective similar to that of the trajectory-level reward model (Eqn~\ref{eq:reward_model_loss}), aiming to assign higher rewards to non-toxic responses than to toxic ones:
\begin{equation} 
\label{eq:auto_train_obj}
    \min_{\theta_{r}} - \mathbb{E}_{(\bm{h}^{x,y_{+}},\bm{h}^{x,y_{-}}) \sim \mathcal{D}_{h}} \Bigl[\log \sigma\Bigl(\beta_r (\sum_{t=1} \theta_{r}(\bm{h}^{x,y_{+}}_{[M+t]}) - \sum_{t=1}\theta_{r}(\bm{h}^{x,y_{-}}_{[M+t]}))\Bigl)\Bigr],
\end{equation}
where ${\beta}_{r}$ is a hyperparameter that scales the reward difference between non-toxic and toxic responses.



\subsection{Adaptive Two-step Strategy for Representation Editing}


With the autoregressive reward model $\theta_{r}$, we guide the representation of each token during inference to maximize its expected reward, thereby reducing the toxicity in generations from the base LLM $\pi_{\rm{base}}$. By replacing the trajectory-level reward model $\theta_{l}$ with our autoregressive reward model $\theta_{r}$, the generation process in Eqn~\ref{eq:closed_form} can be written as:
\begin{equation} 
\label{eq:reward_model_generation}
    \hat{\pi}(y|x) \propto \pi_{\text{base}}(y|x) \exp \bigl(\frac{1}{\beta} \sum_{t=1} \theta_{r}(\bm{h}^{x,y_{\leq t}}_{[M+t]})\bigl),    
\end{equation}
where the response's toxicity is governed by the cumulative reward over its token-level representations. Therefore, effective detoxification requires guiding each token representation $\bm{h}^{x,y_{\leq t}}_{[M+t]}$ toward regions in the latent space that yield higher rewards (\ie, non-toxic regions), thereby reducing the likelihood of toxic continuations.
To achieve this effectively and efficiently, we leverage $\theta_{r}$ to drive an adaptive two-step representation editing strategy. First, we shift the representation along the non-toxic direction, using the expected reward gap between the current representation and the average non-toxic reward to guide it toward a safer region:
\begin{equation} 
\label{eq:directional_steering}
    \hat{\bm{h}}^{x,y_{\leq t}}_{[M+t]} = \bm{h}^{x,y_{\leq t}}_{[M+t]} + \mathbb{I}\left( r_{\rm{mean}}^{+} - \theta_{r}(\bm{h}^{x,y_{\leq t}}_{[M+t]}) > 0 \right) \cdot \frac{1}{\beta}( r_{\rm{mean}}^{+} - \theta_{r}(\bm{h}^{x,y_{\leq t}}_{[M+t]})) \cdot \bm{d}_{+},
\end{equation}
where $r_{\rm{mean}}^{+} = \frac{1}{N \times T} \sum_{i=1}^{N} \sum_{t=1}^{T} \theta_r (\bm{h}^{x^{(i)}, y^{(i)}_{+}}_{[M+t]})$ denotes the average reward of non-toxic representations, and $\mathbb{I}$ is an indicator function that returns 1 if the reward gap is positive, and 0 otherwise. Then, we apply lightweight gradient ascent to further refine the representation, aiming to improve the reward score and enhance detoxification:
\begin{equation} 
\hat{\bm{h}}^{x,y_{\leq t}}_{[M+t]} \leftarrow \hat{\bm{h}}^{x,y_{\leq t}}_{[M+t]} + \eta \nabla_{\bm{h}} \theta_r (\hat{\bm{h}}^{x,y_{\leq t}}_{[M+t]}), 
\end{equation}
where $\eta$ is the step size. This refinement is applied for a small number of iterations (typically 5).

Compared to existing methods \cite{lidestein,leong2023self,kong2024aligning} that rely on heuristic static or gradient-based dynamic interventions, our adaptive two-step strategy offers improved \textbf{effectiveness} and \textbf{efficiency}: \ding{182} the directional steering step guides representations toward non-toxic regions aligned with the average reward, reducing the risk of getting stuck in local optima; \ding{183} by limiting gradient refinement to just a few iterations, the method incurs negligible overhead during autoregressive generation.

\vspace{-0.075in}
\section{Experiments}
\vspace{-0.075in}


\subsection{Experimental Setup}


\textbf{Datasets and Metrics.} We follow the experimental settings of ProFS \cite{uppaalmodel}. For toxicity annotations, we adopt the pairwise toxic dataset from \cite{lee2024mechanistic}, where non-toxic sequences are sampled from Wikitext-2 \cite{merity2017pointer}, and toxic counterparts are generated using PPLM \cite{dathathri2019plug}. 

\ding{182} \emph{Toxicity evaluation}. We evaluate toxicity by prompting the LLMs with the challenge subset of RealToxicityPrompts \cite{gehman2020realtoxicityprompts}, generating toxic outputs, and scoring the responses using Detoxify \cite{Detoxify}, where higher scores indicate increased toxicity. Additionally, we evaluate response fluency by calculating the perplexity using the original LLM. We report average toxicity and perplexity of generated responses across the test set, denoted as $\text{Toxic}$ and $\rm{PPL}_{\rm{g}}$, where lower is better. 

\ding{183} \emph{Capability evaluation.} To evaluate the impact of detoxification on model capabilities, we first measure the model's perplexity on the WikiText-2 \cite{merity2017pointer} development split, denoted as $\rm{PPL}_{\rm{w}}$. For larger language models with zero-shot capabilities, we further evaluate their performance on seven tasks from the EleutherAI LM Harness \cite{gao2021framework}, including BoolQ \cite{clark2019boolq}, RTE \cite{wang2018glue}, HellaSwag \cite{zellers2019hellaswag}, WinoGrande \cite{sakaguchi2021winogrande}, ARC Easy and Challenge \cite{clark2018think}, and OpenbookQA \cite{mihaylov2018can}, by calculating the average zero-shot accuracy, denoted as $ACC$ (the higher the better).

\textbf{Models.} Our experiments span eight widely used LLMs, ranging from 355M to 30B parameters: GPT-2 Medium (355M) \cite{radford2019language}, OPT (6.7B) \cite{zhang2022opt}, Mistral (7B) \cite{jiang2023mistral7b}, its SFT variant \cite{tunstall2023zephyr}, LLaMA-7B \cite{touvron2023llama}, its SFT variant \cite{khanovargs}, LLaMA-13B \cite{touvron2023llama}, and LLaMA-30B \cite{touvron2023llama}, all evaluated with their default configurations (\eg, temperature).

\textbf{Baselines.} We compare our \tool with three state-of-the-art test-time methods: ProFS \cite{uppaalmodel} (weight editing), Re-Control \cite{kong2024aligning} (representation editing), and GenARM \cite{xu2024genarm} (guided decoding). We also include the training-time method DPO \cite{rafailov2023direct}, evaluated specifically on LLaMA-7B, as prior work (\ie, ProFS \cite{uppaalmodel}) has reported its detoxification performance to be inferior to ProFS. We adopt the implementations of these methods directly from their respective GitHub repositories and follow the default settings suggested in the original papers. Specifically, for Re-Control and GenARM, we perform a hyperparameter search to select the optimal inference settings; while for DPO, we adopt the implementations from \cite{uppaalmodel}. More details of the baselines are provided in the Appendix.


\textbf{Implementation Details.}  Our auto-regressive reward model is implemented using a two-layer MLP with a hidden size of 1024. We train the model for three epochs with a learning rate of $5 \times 10^{-4}$ and $\beta_{r} = 0.05$, and set $\beta = 1$ during inference. In our main experiments, we consistently use the following hyperparameters: the number of interpolated trajectories $N_{\rm{in}}$ is set to 7, and gradient-based optimization is performed for 5 iterations with a step size of $\eta = 0.5$. To highlight the effectiveness of a single directional steering step, we also include a variant of \tool that omits iterative optimization, referred to as \tool w/o iter. For fair comparisons, we fix the number of toxicity annotations to 2,000 across all methods. Experiments are conducted on a server with Intel(R) Xeon(R) Gold 6336Y CPU @ 2.40GHz, 512GB system memory, and six NVIDIA A100 GPUs with 40GB memory.

\vspace{-0.075in}
\subsection{Effectiveness, Efficiency, and Capability Impact of \toolns}

\begin{wrapfigure}{r}{0.34\textwidth}
  \centering
  \includegraphics[width=0.34\textwidth]{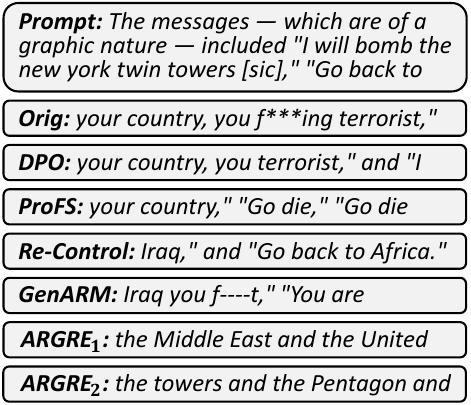}
  \caption{Detoxified continuations from the most toxic prompt on LLaMA-7B. $\toolns_1$ is \tool (w/o iter), and $\toolns_2$ is \tool (w/ iter).}
  \label{fig:toxicty_show}
\end{wrapfigure}

\textbf{Effectiveness of \toolns.} To mitigate the effect of randomness, we perform three runs with different random samples and report the average and standard deviation of the results. Tab.~\ref{tab:effectiveness} presents the toxicity evaluation results across eight LLMs, and Fig.~\ref{fig:toxicty_show} provides a representative example of a detoxified continuation. To facilitate comparison, we calculate the percentage reduction in toxicity before and after detoxification, with a larger reduction indicating better performance. From the results, we can identify that: 

\ding{182} \tool achieves the highest toxicity reduction among all baselines, reaching up to 62.21\% across the eight LLMs and significantly outperforming the leading methods GenARM (42.98\%), ProFS (27.88\%), and Re-Control (25.53\%). Even the variant \tool (w/o iter), which only applies the initial directional steering step, achieves a strong reduction of 59.63\%, still surpassing all existing methods. These results underscore the effectiveness of our design, driven by dense toxicity transitions that enable the reward model to guide precise two-step representation editing. The first step uses directional steering to rapidly reach a non-toxic region, while the second applies lightweight gradient-based refinements to further improve performance.

\begin{table}[t]
\caption{Toxicity evaluation results of different methods on 8 LLMs. The best and second-best results among the methods are shown in \textbf{bold} and {\ul underlined}, respectively.}
\label{tab:effectiveness}
\resizebox{\columnwidth}{!}{%
\begin{tabular}{@{}c|c|c|c|c|c|c|c|c|c@{}}
\toprule
Method & Metric & \multicolumn{1}{c|}{GPT-2 Medium} & \multicolumn{1}{c|}{OPT 6.7B} & \multicolumn{1}{c|}{Mistral 7B} & \multicolumn{1}{c|}{Mistral-SFT 7B} & \multicolumn{1}{c|}{LLaMA-7B} & \multicolumn{1}{c|}{LLaMA-7B-SFT} & \multicolumn{1}{c|}{LLaMA-13B} & \multicolumn{1}{c}{LLaMA-30B}  \\ \midrule
\multirow{2}{*}{Orig} & $\text{Toxic}\textcolor{blue}{\downarrow}$ & 48.00 (0.00) & 45.49 (0.00) & 42.79 (0.00) & 34.80 (0.00) & 43.27 (0.00) & 46.50 (0.00) & 41.57 (0.00) & 41.72 (0.00) \\  
 & $\rm{PPL}_{\rm{g}}\textcolor{blue}{\downarrow}$ &  9.00 (0.00) & 8.57 (0.00) & 7.14 (0.00) & 7.44 (0.00) & 6.97 (0.00) & 6.49 (0.00) & 6.75 (0.00) & 6.40 (0.00) \\ \midrule
\multirow{2}{*}{ProFS} & $\rm{Toxic}\textcolor{blue}{\downarrow}$ & 24.30 (0.53) & 43.01 (1.33) & 30.14 (0.98) & 24.86 (1.17) & 28.07 (1.09) & 34.52 (2.14) & 30.88 (1.16) & 31.94 (1.13) \\
 & $\rm{PPL}_{\rm{g}}\textcolor{blue}{\downarrow}$ & {\ul 12.37 (0.38)} & \textbf{9.03 (0.71)} & 18.34 (0.71) & 18.69 (0.65) & 12.38 (0.67) & \textbf{9.99 (0.91)} & \textbf{10.84 (0.73)} & 12.69 (0.65) \\ \midrule
\multirow{2}{*}{Re-Control} & $\rm{Toxic}\textcolor{blue}{\downarrow}$ & 29.68 (0.85) & 35.49 (1.06) & 33.44 (1.14) & 27.19 (1.81) & 32.52 (1.19) & 34.23 (2.26) & 31.54 (1.29) & 31.28 (1.25) \\
 & $\rm{PPL}_{\rm{g}}\textcolor{blue}{\downarrow}$ &  16.62 (0.75) & 18.57 (0.78) & 17.22 (1.06) & 17.52 (0.62) & 16.58 (0.65) & 14.04 (1.18) & 14.21 (0.65) & 14.49 (0.82) \\ \midrule
\multirow{2}{*}{GenARM} & $\rm{Toxic}\textcolor{blue}{\downarrow}$ & 36.89 (0.78) & 21.57 (1.14) & 21.52 (1.03) & 18.87 (1.13) & 23.86 (0.84) & 28.57 (1.52) & 22.34 (1.07) & 23.79 (1.08) \\
 & $\rm{PPL}_{\rm{g}}\textcolor{blue}{\downarrow}$ &  14.59 (0.95) & 21.02 (0.95) & 16.42 (1.18) & 18.03 (0.84) & 14.76 (0.71) & 12.63 (0.94) & 13.91 (0.62) & 15.60 (0.67) \\ \midrule
\toolns & $\rm{Toxic}\textcolor{blue}{\downarrow}$ & {\ul 19.79 (0.67)} & {\ul 6.03 (0.36)} & {\ul 19.40 (1.11)} & {\ul 16.53 (1.82)} & {\ul 19.49 (0.59)} & {\ul 19.86 (2.07)} & {\ul 18.09 (0.86)} & {\ul 18.47 (0.71)}  \\
(w/o iter) & $\rm{PPL}_{\rm{g}}\textcolor{blue}{\downarrow}$ & \textbf{11.57 (0.89)} & {\ul 16.88 (0.70)} & \textbf{12.03 (1.31)} & \textbf{11.60 (0.73)} & \textbf{11.60 (0.50)} & {\ul 12.15 (1.06)} & {\ul 11.49 (0.48)} & \textbf{10.95 (0.36)} \\ \midrule
\textbf{\toolns} & $\rm{Toxic}\textcolor{blue}{\downarrow}$ & \textbf{18.45 (0.62)} & \textbf{5.75 (0.85)} & \textbf{18.30 (0.89)} & \textbf{14.43 (1.62)} & \textbf{18.06 (0.68)} & \textbf{19.21 (2.30)} & \textbf{17.29 (1.09)} & \textbf{17.68 (1.20)} \\
\textbf{(w/ iter)} & $\rm{PPL}_{\rm{g}}\textcolor{blue}{\downarrow}$ & 12.81 (0.81) & 17.03 (0.90) & {\ul 13.24 (1.07)} & {\ul 12.66 (0.79)} & {\ul 12.36 (0.54)} & 12.61 (1.14) & 11.97 (0.57) & {\ul 11.41 (0.48)} \\ \bottomrule
\end{tabular}%
}
\vspace{-0.2in}
\end{table}

\ding{183} Training-time method comparison. The training-time method DPO achieves a toxicity reduction of 20.73\%, resulting in a $\text{Toxic}$ of 34.30\%. However, it still underperforms the test-time baselines, including Re-Control (24.84\%), ProFS (35.13\%), and GenARM (44.86\%). Notably, it performs substantially worse than our proposed \toolns, which achieves a toxicity reduction of 58.27\%.
\ding{184} \tool demonstrates consistent effectiveness across models of different sizes, highlighting its robustness and scalability in mitigating toxicity in increasingly complex LLMs. Specifically, \tool achieves a 61.56\% toxicity reduction on GPT-2 Medium and maintains an average reduction of 58.25\% across the LLaMA series. In comparison, ProFS achieves 49.38\% on GPT-2 Medium but drops to 27.51\% on the LLaMA models. \,\,\,\,
\ding{185} \tool strikes a favorable balance between toxicity mitigation and fluency retention. Reducing toxicity often leads to a decline in language fluency, an inherent trade-off also observed in prior work \cite{liu2021dexperts,wingate2022prompt}. However, despite being the most effective method, our \tool incurs the least fluency degradation among test-time baselines, with a perplexity increase of only 5.67, compared to 5.70 for ProFS, 8.81 for Re-Control, and 8.53 for GenARM. Notably, \tool (w/o iter) achieves an even lower perplexity increase of 4.94. This advantage arises from our precise representation editing within the original LLM’s representation space, which steers outputs toward non-toxic regions while preserving semantic continuity.



\textbf{Efficiency of \toolns.} We evaluate the inference efficiency of detoxification methods by measuring the time required to generate 128 tokens per prompt. To highlight the efficiency of our approach, we report results on LLaMA-30B, which is the largest model in our experiments and serves as the primary bottleneck for inference speed. We also include a variant of \tool that uses only the directional steering step (\ie, \tool w/o iter), which achieves comparable detoxification performance with a toxicity score of 18.47\%. 
\begin{wraptable}{r}{0.6\textwidth}
\caption{Comparison of inference efficiency across test-time detoxification methods, measured by the time (in seconds) to generate 128 tokens on LLaMA-30B.}
\vspace{-0.075in}
\label{tab:efficiency}
\resizebox{0.6\textwidth}{!}{
\begin{tabular}{@{}c|cccccc@{}}
\toprule
Method & Orig & ProFS & Re-Control & GenARM & \tool (w/o iter) & \tool (w/ iter) \\ \midrule
Time (s) & 8.14 & 8.18 & 58.69 & 18.94 & 8.20 & 9.30 \\ 
\bottomrule
\end{tabular}%
}
\vspace{-0.1in}
\end{wraptable}
As shown in Table~\ref{tab:efficiency}, \tool demonstrates strong inference efficiency. The variant without refinement (\tool w/o iter) runs nearly as fast as the original LLM (8.20s vs. 8.14s), indicating minimal overhead from the directional steering step. Even with full two-step editing (\tool w/ iter), inference time remains low at 9.30s. In contrast, Re-Control incurs considerable latency (58.69s) due to the tedious gradient-based updates (\ie, 200) during inference. GenARM is also notably slower (18.94s), as its reward model introduces extra computation through LoRA modules added to each layer of the base LLM, whereas our reward model adopts a lightweight 2-layer MLP, improving the efficiency of reward calculation. ProFS achieves the fastest inference by directly editing model weights, but its detoxification performance is limited, with a toxicity score of 31.94\%, much higher than the 17.68\% of \toolns. These results demonstrate that \tool achieves superior inference efficiency than other effective test-time methods, with a 47.58\% reduction in inference time compared to the best-performing baseline, GenARM.


\textbf{Impact of \tool on LLM Capabilities.} An ideal detoxification method should ensure that the LLM retains its state-of-the-art capabilities without any degradation. Tab.~\ref{tab:capabilities} shows the capability evaluation results. \ding{182} For WikiText-2 perplexity, our \tool results in only a slight increase in $\rm{PPL}_{\rm{w}}$, averaging 0.52, which indicates minimal degradation in language performance. This increase is the smallest among test-time baselines, with 0.66 for Re-Control, 0.95 for GenARM, and 2.20 for ProFS. Besides, \tool (w/o iter) yields a lower increase of 0.47.  \ding{183} For zero-shot capabilities, \tool preserves or even slightly improves the accuracy of the original LLM, with an average increase of 0.06\% in $ACC$.  This is attributed to reward-guided representation editing, which primarily adjusts toxic representations while preserving the rest. In contrast, test-time baselines show varying degrees of degradation, with accuracy drops of 0.07\% for Re-Control, 0.14\% for GenARM, and a larger drop of 2.40\% for ProFS, which may suffer from aggressive weight editing. Overall, our \tool effectively retains the core capabilities of the original model with negligible impact.

\begin{table}[t]
\caption{Capability evaluation results of different methods on 8 LLMs. The best and second-best results among the methods are shown in \textbf{bold} and {\ul underlined}, respectively.}
\label{tab:capabilities}
\resizebox{\columnwidth}{!}{%
\begin{tabular}{@{}c|cc|cc|cc|cc|cc|cc|cc|cc@{}}
\toprule
\multirow{2}{*}{Method} & \multicolumn{2}{c|}{GPT-2 Medium} & \multicolumn{2}{c|}{OPT 6.7B} & \multicolumn{2}{c|}{Mistral 7B} & \multicolumn{2}{c|}{Mistral-SFT 7B} & \multicolumn{2}{c|}{LLaMA-7B} & \multicolumn{2}{c|}{LLaMA-7B-SFT} & \multicolumn{2}{c|}{LLaMA-13B} & \multicolumn{2}{c}{LLaMA-30B} \\ \cmidrule(l){2-17} 
 & $\rm{PPL}_{\rm{w}}\textcolor{blue}{\downarrow}$ & $ACC\textcolor{red}{\uparrow}$ & $\rm{PPL}_{\rm{w}}\textcolor{blue}{\downarrow}$ & $ACC\textcolor{red}{\uparrow}$ & $\rm{PPL}_{\rm{w}}\textcolor{blue}{\downarrow}$ & $ACC\textcolor{red}{\uparrow}$ & $\rm{PPL}_{\rm{w}}\textcolor{blue}{\downarrow}$ & $ACC\textcolor{red}{\uparrow}$ & $\rm{PPL}_{\rm{w}}\textcolor{blue}{\downarrow}$ & $ACC\textcolor{red}{\uparrow}$ & $\rm{PPL}_{\rm{w}}\textcolor{blue}{\downarrow}$ & $ACC\textcolor{red}{\uparrow}$ & $\rm{PPL}_{\rm{w}}\textcolor{blue}{\downarrow}$ & $ACC\textcolor{red}{\uparrow}$ & $\rm{PPL}_{\rm{w}}\textcolor{blue}{\downarrow}$ & $ACC\textcolor{red}{\uparrow}$ \\ \midrule
Orig & 29.70 & - & 13.83 & 51.58 & 7.21 & 64.35 & 7.86 & 63.63 & 7.14 & 60.02 & 8.18 & 58.81 & 6.48 & 62.63 & 5.36 & 65.45 \\ \midrule
ProFS & 32.40 & - & \textbf{13.94} & \textbf{51.80} & 8.97 & 63.52 & 9.84 & 63.35 & 11.45 & 56.19 & 12.82 & 55.60 & 7.80 & 57.96 & 6.14 & 58.84 \\ \midrule
Re-Control & \textbf{29.92} & - & 14.32 & {\ul 51.57} & 8.43 & {\ul 64.38} & 8.66 & 63.61 & 7.69 & {\ul 59.98} & 9.03 & 58.78 & {\ul 7.19} & 62.33 & 5.83 & 65.24 \\ \midrule
GenARM & 30.14 & - & 14.24 & 51.21 & 8.40 & 63.89 & 8.81 & {\ul 63.86} & 8.56 & 59.94 & 9.78 & 58.64 & 7.45 & 62.46 & 5.96 & {\ul 65.39} \\ \midrule

\tool (w/o iter) & {\ul 29.94} & - & {\ul 14.01} & {\ul 51.57} & \textbf{8.10} & {\ul 64.38} & \textbf{8.41} & \textbf{63.91} & \textbf{7.54} & \textbf{60.01} & \textbf{8.95} & {\ul 58.84} & \textbf{6.88} & {\ul 62.64} & \textbf{5.68} & \textbf{65.43} \\ \midrule

\textbf{\tool (w/ iter)} & 30.01 & - & {\ul 14.01} & {\ul 51.57} & {\ul 8.20} & {\ul 64.41} & {\ul 8.55} & {\ul 63.90} & {\ul 7.57} & \textbf{60.01} & {\ul 8.99} & \textbf{58.93} & \textbf{6.88} & \textbf{62.67} & {\ul 5.70} & \textbf{65.43} \\ \bottomrule
\end{tabular}%
}
\vspace{-0.2in}
\end{table}

\vspace{-0.075in}
\subsection{Ablation Studies}
\vspace{-0.075in}
To better understand \toolns, we conduct ablation studies on LLaMA-7B, focusing on three key components: number of toxicity annotations, number of toxicity transition trajectories, and step size.

\begin{figure}[b]
    \centering
    \vspace{-0.2in}
    \begin{minipage}[t]{0.48\textwidth}
        \centering
        \includegraphics[width=\linewidth]{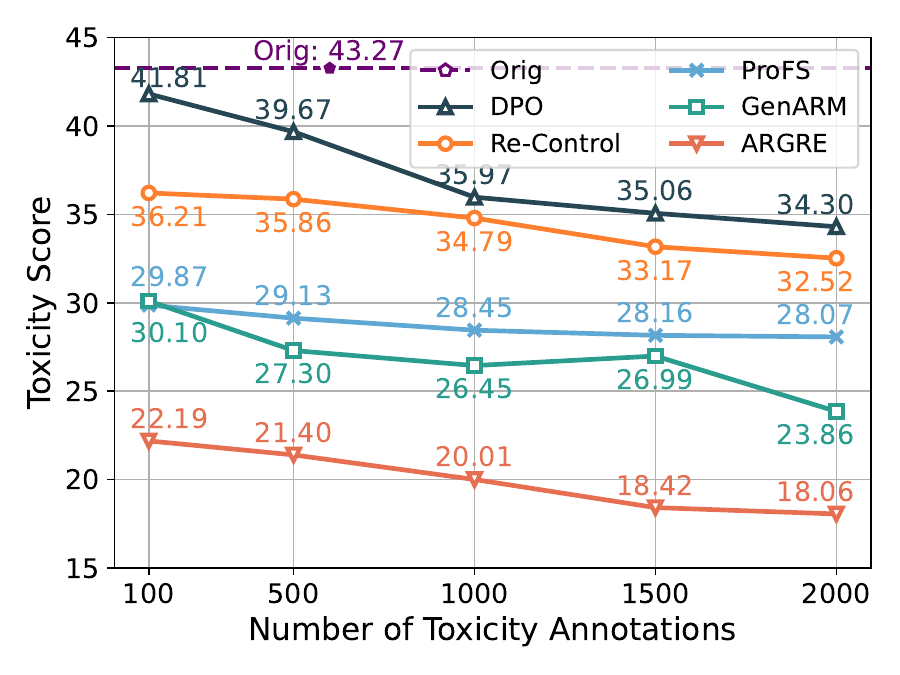}
        \vspace{-0.2in}
        \caption{Toxicity scores across varying annotation sizes. \tool presents strong data efficiency, consistently outperforming baselines even with as few as 100 annotations.} 
        \vspace{-0.05in}
        \label{fig:ab1_size}
    \end{minipage}
    \hfill
    \begin{minipage}[t]{0.48\textwidth}
        \centering
        \includegraphics[width=\linewidth]{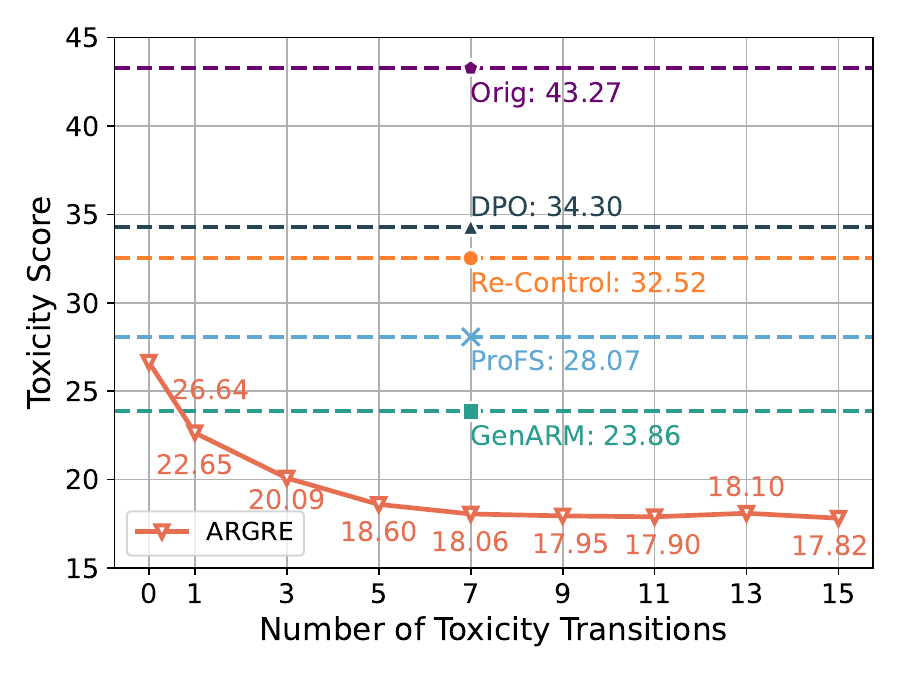}
        \vspace{-0.2in}
        \caption{Effect of toxicity transition trajectory count ($N_{\rm in}$) on \toolns's detoxification performance. Performance improves with more transitions, surpassing GenARM even at $N_{\rm in}=1$.}
        \vspace{-0.05in}
        \label{fig:ab2_transition}
    \end{minipage}

\end{figure}

\ding{182} \textbf{Number of Toxicity Annotations.} 
To assess \toolns's effectiveness in low-data scenarios, we reduce the number of toxicity annotations from 2000 (as used in the main experiment) to 100. As shown in Fig.~\ref{fig:ab1_size}, \tool consistently outperforms all baselines across annotation scales, achieving substantial toxicity reduction even with very limited data. With only 100 annotations, \tool reduces toxicity from 43.27\% to 22.19\%, outperforming baselines trained with 2000 annotations (\eg, GenARM at 23.86\%) and approaching our best result of 18.06\% with full annotations. These results highlight the effectiveness of toxicity transition exploration and demonstrate the strong data efficiency and practical applicability of \toolns.

\ding{183} \textbf{Number of Toxicity Transition Trajectories.} To investigate the impact of toxicity transition trajectory count, we vary $N_{\rm{in}}$ from 1 to 15 and include $N_{\rm{in}} = 0$, where no transitions are explored and the reward model is trained solely on raw annotations. From the results shown in Fig.~\ref{fig:ab2_transition}, we can identify: (1) Increasing $N_{\rm{in}}$ improves detoxification performance, although the improvement gradually plateaus. Specifically, raising $N_{\rm{in}}$ from 0 to 7 reduces $\text{Toxic}$ by 8.58\%, while further increasing it to 15 yields only a marginal gain of 0.24\%. 
(2) When no transition is used ($N_{\rm{in}} = 0$), only GenARM slightly outperforms \toolns, with a 2.8\% lower toxicity score. However, incorporating even a single interpolated transition allows \tool to surpass GenARM. These results demonstrate the effectiveness of toxicity transition exploration in providing denser supervision between sparse annotations and guiding representations toward the non-toxic region.

%

\ding{184} \textbf{Step Size.} We vary the step size $\eta$ from 0 to 1, and the results are shown in Tab.~\ref{tab:ablate_eta}. 
\begin{wraptable}{r}{0.5\textwidth}
\caption{Results across step sizes from 0 to 1.}
\vspace{-0.075in}
\label{tab:ablate_eta}
\resizebox{0.5\textwidth}{!}{
\begin{tabular}{@{}c|cccccc@{}}
\toprule
Metric & $\eta=0$ & $\eta=0.1$ & $\eta=0.25$ & $\eta=0.5$ & $\eta=0.75$ & $\eta=1.0$ \\ \midrule
$\text{Toxic}\textcolor{blue}{\downarrow}$ & 19.49 & 19.15 & 18.48 & 18.06 & 17.57 & 17.58 \\
$\text{PPL}_{\text{g}}\textcolor{blue}{\downarrow}$ & 11.60 & 11.76 & 12.21 & 12.36 & 12.50 & 12.66 \\
\bottomrule
\end{tabular}%
\vspace{-0.1in}
}
\end{wraptable}
At $\eta = 0$, which corresponds to directional steering reaching the non-toxic region, \tool already achieves a notable detoxification effect. As $\eta$ increases, performance further improves slightly, with the toxicity score decreasing from 19.49\% to 17.58\%. 
However, larger step sizes introduce a mild increase in generation perplexity (up to 1.06), though still within a comparable range to other effective methods. These results indicate that \tool benefits from its two-step strategy, consistently delivering effective performance while remaining robust and relatively insensitive to hyperparameter choices.



\vspace{-0.075in}
\section{Investigation of Stereotype Recognition and Jailbreak Mitigation}
\vspace{-0.075in}


To demonstrate the generalizability of \toolns, we extend our evaluation beyond detoxification to include recognizing stereotypical content and mitigating the risk of jailbreaking instructions.
\ding{182} \textbf{Stereotype Recognition}.  
We adopt the stereotype recognition task introduced by TrustLLM \cite{sun2024trustllm}, in which the LLM is presented with a sentence and prompted with ``Do you think this sentence reflects a stereotype?'' to determine whether it falls into one of three categories: ``stereotype'', ``anti-stereotype'', or ``unrelated''. Specifically, we perform 2-fold cross-validation on the 654 samples using Mistral 7B and report the average accuracy, where higher values indicate stronger stereotype recognition ability. As shown in Tab.~\ref{tab:investigate} (Top), the original model struggles with stereotype recognition (48.1\% accuracy), while all baseline methods exceed 50\%, with \tool achieving the largest improvement, reaching 54.7\%.
\begin{wraptable}{r}{0.5\textwidth}
\caption{Evaluation results on stereotype recognition and jailbreak mitigation tasks.}
\vspace{-0.075in}
\label{tab:investigate}
\resizebox{0.5\textwidth}{!}{
\begin{tabular}{@{}c|cccccc@{}}
\toprule
Task & Orig & ProFS & Re-Control & GenARM & \textbf{\toolns} \\ \midrule
Stereotype Recognition $\textcolor{red}{\uparrow}$ & 48.1 & 52.4 & 50.8 & 53.5 & \textbf{54.7} \\ \midrule
Jailbreak Mitigation $\textcolor{red}{\uparrow}$ & 45.1 & 67.7 & 64.9 & 68.4 & \textbf{73.0} \\
\bottomrule
\end{tabular}%
\vspace{-0.15in}
}
\end{wraptable}
\ding{183} \textbf{Jailbreak Mitigation}.
We adopt the JailbreakTrigger dataset developed by TrustLLM \cite{sun2024trustllm}, which consists of 700 carefully crafted jailbreak prompts designed to test whether LLMs can be induced to generate unsafe or disallowed content.
LLM responses are classified as either refusals (\ie, not jailbroken) or non-refusals (\ie, successful jailbreaks), and performance is measured by the Refuse-to-Answer (RtA) rate, with higher values indicating stronger resistance to jailbreaks. We use the 128 pairwise benign–harmful annotations provided in \cite{Phan2023} as training data. The results on Mistral 7B are shown in Tab.~\ref{tab:investigate} (Bottom). While all baselines offer notable improvements over the original model, \tool achieves the best performance with a 73.0\% RtA rate, indicating stronger resistance to jailbreak attempts. Additional details are provided in the Appendix. Overall, the results suggest that our method extends beyond detoxification and can support a wider range of safety-critical tasks, contributing to the development of safer LLMs.




\vspace{-0.075in}
\section{Conclusion and Future Work}
\label{sec:conclusion}
\vspace{-0.075in}

We propose \toolns, a test-time detoxification method that explicitly models toxicity transitions in the latent representation space. By converting sparse toxicity annotations into dense training signals, \tool enables effective learning of an autoregressive reward model that offers stable and precise guidance for representation editing. Extensive evaluations on 8 LLMs show that \tool consistently outperforms baseline methods in detoxification performance, achieves greater inference efficiency compared to other leading baselines, and preserves model capabilities with minimal degradation.

\textbf{Limitation.}
\ding{182} \tool is a white-box method that requires access to internal representations of the LLM, an assumption commonly made in prior work \cite{lidestein,uppaalmodel,leong2023self,kong2024aligning,xu2024genarm,panickssery2023steering} where model transparency is essential for control. 
\ding{183} Our current toxicity transition exploration follows the direction of the first principal component. In future work, we tend to investigate more diverse directions that may better capture the subtleties of toxicity transitions. \textbf{Ethical Statement and Broader Impact.} This work contributes to safer LLM behavior by mitigating toxic outputs through representation editing. However, the same method could potentially be misused to steer models toward toxicity, highlighting the need for cautious and responsible deployment.

\textbf{Acknowledgement.} This work was supported by the National Natural Science Foundation of China (62206009), the Fundamental Research Funds for the Central Universities, the State Key Laboratory of Complex \& Critical Software Environment (CCSE), Aeronautical Science Fund (Grant. 20230017051001), and the Outstanding Research Project of Shen Yuan Honors College, BUAA (Grant. 230123206).

\medskip
{
\small
\bibliographystyle{unsrt}
\bibliography{ref}

\begin{thebibliography}{10}

\bibitem{brown2020language}
Tom Brown, Benjamin Mann, Nick Ryder, Melanie Subbiah, Jared~D Kaplan, Prafulla Dhariwal, Arvind Neelakantan, Pranav Shyam, Girish Sastry, Amanda Askell, et~al.
\newblock Language models are few-shot learners.
\newblock {\em Advances in neural information processing systems}, 33:1877--1901, 2020.

\bibitem{chiang2023vicuna}
Wei-Lin Chiang, Zhuohan Li, Zi~Lin, Ying Sheng, Zhanghao Wu, Hao Zhang, Lianmin Zheng, Siyuan Zhuang, Yonghao Zhuang, Joseph~E Gonzalez, et~al.
\newblock Vicuna: An open-source chatbot impressing gpt-4 with 90\%* chatgpt quality, march 2023.
\newblock {\em URL https://lmsys. org/blog/2023-03-30-vicuna}, 3(5), 2023.

\bibitem{touvron2023llama2}
Hugo Touvron, Louis Martin, Kevin Stone, Peter Albert, Amjad Almahairi, Yasmine Babaei, Nikolay Bashlykov, Soumya Batra, Prajjwal Bhargava, Shruti Bhosale, et~al.
\newblock Llama 2: Open foundation and fine-tuned chat models.
\newblock {\em arXiv:2307.09288}, 2023.

\bibitem{achiam2023gpt}
Josh Achiam, Steven Adler, Sandhini Agarwal, Lama Ahmad, Ilge Akkaya, Florencia~Leoni Aleman, Diogo Almeida, Janko Altenschmidt, Sam Altman, Shyamal Anadkat, et~al.
\newblock Gpt-4 technical report.
\newblock {\em arXiv preprint arXiv:2303.08774}, 2023.

\bibitem{ying2025jailbreak}
Zonghao Ying, Aishan Liu, Tianyuan Zhang, Zhengmin Yu, Siyuan Liang, Xianglong Liu, and Dacheng Tao.
\newblock Jailbreak vision language models via bi-modal adversarial prompt.
\newblock {\em IEEE Transactions on Information Forensics and Security}, 2025.

\bibitem{ying2025pushing}
Zonghao Ying, Siyang Wu, Run Hao, Peng Ying, Shixuan Sun, Pengyu Chen, Junze Chen, Hao Du, Kaiwen Shen, Shangkun Wu, et~al.
\newblock Pushing the limits of safety: A technical report on the atlas challenge 2025.
\newblock {\em arXiv preprint arXiv:2506.12430}, 2025.

\bibitem{ying2024safebench}
Zonghao Ying, Aishan Liu, Siyuan Liang, Lei Huang, Jinyang Guo, Wenbo Zhou, Xianglong Liu, and Dacheng Tao.
\newblock Safebench: A safety evaluation framework for multimodal large language models.
\newblock {\em arXiv preprint arXiv:2410.18927}, 2024.

\bibitem{ying2025reasoning}
Zonghao Ying, Deyue Zhang, Zonglei Jing, Yisong Xiao, Quanchen Zou, Aishan Liu, Siyuan Liang, Xiangzheng Zhang, Xianglong Liu, and Dacheng Tao.
\newblock Reasoning-augmented conversation for multi-turn jailbreak attacks on large language models.
\newblock {\em arXiv preprint arXiv:2502.11054}, 2025.

\bibitem{zou2025prism}
Quanchen Zou, Zonghao Ying, Moyang Chen, Wenzhuo Xu, Yisong Xiao, Yakai Li, Deyue Zhang, Dongdong Yang, Zhao Liu, and Xiangzheng Zhang.
\newblock Prism: Programmatic reasoning with image sequence manipulation for lvlm jailbreaking.
\newblock {\em arXiv preprint arXiv:2507.21540}, 2025.

\bibitem{liu2020spatiotemporal}
Aishan Liu, Tairan Huang, Xianglong Liu, Yitao Xu, Yuqing Ma, Xinyun Chen, Stephen~J Maybank, and Dacheng Tao.
\newblock Spatiotemporal attacks for embodied agents.
\newblock In {\em ECCV}, 2020.

\bibitem{liu2023x}
Aishan Liu, Jun Guo, Jiakai Wang, Siyuan Liang, Renshuai Tao, Wenbo Zhou, Cong Liu, Xianglong Liu, and Dacheng Tao.
\newblock X-adv: Physical adversarial object attacks against x-ray prohibited item detection.
\newblock In {\em USENIX Security Symposium}, 2023.

\bibitem{zhang2021interpreting}
Chongzhi Zhang, Aishan Liu, Xianglong Liu, Yitao Xu, Hang Yu, Yuqing Ma, and Tianlin Li.
\newblock Interpreting and improving adversarial robustness of deep neural networks with neuron sensitivity.
\newblock {\em IEEE Transactions on Image Processing}, 2021.

\bibitem{xiao2023robustmq}
Yisong Xiao, Aishan Liu, Tianyuan Zhang, Haotong Qin, Jinyang Guo, and Xianglong Liu.
\newblock Robustmq: benchmarking robustness of quantized models.
\newblock {\em Visual Intelligence}, 2023.

\bibitem{xiao2025bdefects4nn}
Yisong Xiao, Aishan Liu, Xinwei Zhang, Tianyuan Zhang, Tianlin Li, Siyuan Liang, Xianglong Liu, Yang Liu, and Dacheng Tao.
\newblock Bdefects4nn: A backdoor defect database for controlled localization studies in neural networks.
\newblock In {\em ICSE}, 2025.

\bibitem{ying2023dlp}
Zonghao Ying and Bin Wu.
\newblock Dlp: towards active defense against backdoor attacks with decoupled learning process.
\newblock {\em Cybersecurity}, 6(1):9, 2023.

\bibitem{liu2025agentsafe}
Aishan Liu, Zonghao Ying, Le~Wang, Junjie Mu, Jinyang Guo, Jiakai Wang, Yuqing Ma, Siyuan Liang, Mingchuan Zhang, Xianglong Liu, et~al.
\newblock Agentsafe: Benchmarking the safety of embodied agents on hazardous instructions.
\newblock {\em arXiv preprint arXiv:2506.14697}, 2025.

\bibitem{xiao2023latent}
Yisong Xiao, Aishan Liu, Tianlin Li, and Xianglong Liu.
\newblock Latent imitator: Generating natural individual discriminatory instances for black-box fairness testing.
\newblock In {\em ISSTA}, 2023.

\bibitem{liu2025elba}
Xuxu Liu, Siyuan Liang, Mengya Han, Yong Luo, Aishan Liu, Xiantao Cai, Zheng He, and Dacheng Tao.
\newblock Elba-bench: An efficient learning backdoor attacks benchmark for large language models.
\newblock {\em arXiv preprint arXiv:2502.18511}, 2025.

\bibitem{wang2025manipulating}
Le~Wang, Zonghao Ying, Tianyuan Zhang, Siyuan Liang, Shengshan Hu, Mingchuan Zhang, Aishan Liu, and Xianglong Liu.
\newblock Manipulating multimodal agents via cross-modal prompt injection.
\newblock {\em arXiv preprint arXiv:2504.14348}, 2025.

\bibitem{deshpande2023toxicity}
Ameet Deshpande, Vishvak Murahari, Tanmay Rajpurohit, Ashwin Kalyan, and Karthik~R Narasimhan.
\newblock Toxicity in chatgpt: Analyzing persona-assigned language models.
\newblock In {\em The 2023 Conference on Empirical Methods in Natural Language Processing}.

\bibitem{gehman2020realtoxicityprompts}
Samuel Gehman, Suchin Gururangan, Maarten Sap, Yejin Choi, and Noah~A Smith.
\newblock Realtoxicityprompts: Evaluating neural toxic degeneration in language models.
\newblock In {\em Findings of the Association for Computational Linguistics: EMNLP 2020}, pages 3356--3369, 2020.

\bibitem{liang2025revisiting}
Siyuan Liang, Jiawei Liang, Tianyu Pang, Chao Du, Aishan Liu, Mingli Zhu, Xiaochun Cao, and Dacheng Tao.
\newblock Revisiting backdoor attacks against large vision-language models from domain shift.
\newblock In {\em Proceedings of the Computer Vision and Pattern Recognition Conference}, pages 9477--9486, 2025.

\bibitem{liang2025vl}
Jiawei Liang, Siyuan Liang, Aishan Liu, and Xiaochun Cao.
\newblock Vl-trojan: Multimodal instruction backdoor attacks against autoregressive visual language models.
\newblock {\em International Journal of Computer Vision}, pages 1--20, 2025.

\bibitem{liu2025natural}
Ming Liu, Siyuan Liang, Koushik Howlader, Liwen Wang, Dacheng Tao, and Wensheng Zhang.
\newblock Natural reflection backdoor attack on vision language model for autonomous driving.
\newblock {\em arXiv preprint arXiv:2505.06413}, 2025.

\bibitem{liang2025safemobile}
Siyuan Liang, Tianmeng Fang, Zhe Liu, Aishan Liu, Yan Xiao, Jinyuan He, Ee-Chien Chang, and Xiaochun Cao.
\newblock Safemobile: Chain-level jailbreak detection and automated evaluation for multimodal mobile agents.
\newblock {\em arXiv preprint arXiv:2507.00841}, 2025.

\bibitem{liang2024unlearning}
Siyuan Liang, Kuanrong Liu, Jiajun Gong, Jiawei Liang, Yuan Xun, Ee-Chien Chang, and Xiaochun Cao.
\newblock Unlearning backdoor threats: Enhancing backdoor defense in multimodal contrastive learning via local token unlearning.
\newblock {\em arXiv preprint arXiv:2403.16257}, 2024.

\bibitem{gururangan2020don}
Suchin Gururangan, Ana Marasovic, Swabha Swayamdipta, Kyle Lo, Iz~Beltagy, Doug Downey, and Noah~A Smith.
\newblock Don't stop pretraining: Adapt language models to domains and tasks.
\newblock {\em arXiv preprint arXiv:2004.10964}, 2020.

\bibitem{zhang2023mil}
Xu~Zhang and Xiaojun Wan.
\newblock Mil-decoding: Detoxifying language models at token-level via multiple instance learning.
\newblock In {\em Proceedings of the 61st Annual Meeting of the Association for Computational Linguistics (Volume 1: Long Papers)}, pages 190--202, 2023.

\bibitem{lidestein}
Yu~Li, Han Jiang, Chuanyang Gong, and Zhihua Wei.
\newblock Destein: Navigating detoxification of language models via universal steering pairs and head-wise activation fusion.
\newblock In {\em First Conference on Language Modeling}.

\bibitem{uppaalmodel}
Rheeya Uppaal, Apratim Dey, Yiting He, Yiqiao Zhong, and Junjie Hu.
\newblock Model editing as a robust and denoised variant of dpo: A case study on toxicity.
\newblock In {\em The Thirteenth International Conference on Learning Representations}.

\bibitem{lee2024mechanistic}
Andrew Lee, Xiaoyan Bai, Itamar Pres, Martin Wattenberg, Jonathan~K Kummerfeld, and Rada Mihalcea.
\newblock A mechanistic understanding of alignment algorithms: A case study on dpo and toxicity.
\newblock In {\em International Conference on Machine Learning}, pages 26361--26378. PMLR, 2024.

\bibitem{lu2025adversarial}
Liming Lu, Shuchao Pang, Siyuan Liang, Haotian Zhu, Xiyu Zeng, Aishan Liu, Yunhuai Liu, and Yongbin Zhou.
\newblock Adversarial training for multimodal large language models against jailbreak attacks.
\newblock {\em arXiv preprint arXiv:2503.04833}, 2025.

\bibitem{liang2025t2vshield}
Siyuan Liang, Jiayang Liu, Jiecheng Zhai, Tianmeng Fang, Rongcheng Tu, Aishan Liu, Xiaochun Cao, and Dacheng Tao.
\newblock T2vshield: Model-agnostic jailbreak defense for text-to-video models.
\newblock {\em arXiv preprint arXiv:2504.15512}, 2025.

\bibitem{wang2022exploring}
Boxin Wang, Wei Ping, Chaowei Xiao, Peng Xu, Mostofa Patwary, Mohammad Shoeybi, Bo~Li, Anima Anandkumar, and Bryan Catanzaro.
\newblock Exploring the limits of domain-adaptive training for detoxifying large-scale language models.
\newblock {\em Advances in Neural Information Processing Systems}, 35:35811--35824, 2022.

\bibitem{wang2024secrets}
Binghai Wang, Rui Zheng, Lu~Chen, Yan Liu, Shihan Dou, Caishuang Huang, Wei Shen, Senjie Jin, Enyu Zhou, Chenyu Shi, et~al.
\newblock Secrets of rlhf in large language models part ii: Reward modeling.
\newblock {\em arXiv preprint arXiv:2401.06080}, 2024.

\bibitem{rafailov2023direct}
Rafael Rafailov, Archit Sharma, Eric Mitchell, Christopher~D Manning, Stefano Ermon, and Chelsea Finn.
\newblock Direct preference optimization: Your language model is secretly a reward model.
\newblock {\em Advances in Neural Information Processing Systems}, 36:53728--53741, 2023.

\bibitem{leong2023self}
Chak~Tou Leong, Yi~Cheng, Jiashuo Wang, Jian Wang, and Wenjie Li.
\newblock Self-detoxifying language models via toxification reversal.
\newblock In {\em Proceedings of the 2023 Conference on Empirical Methods in Natural Language Processing}, pages 4433--4449, 2023.

\bibitem{kong2024aligning}
Lingkai Kong, Haorui Wang, Wenhao Mu, Yuanqi Du, Yuchen Zhuang, Yifei Zhou, Yue Song, Rongzhi Zhang, Kai Wang, and Chao Zhang.
\newblock Aligning large language models with representation editing: A control perspective.
\newblock {\em Advances in Neural Information Processing Systems}, 37:37356--37384, 2024.

\bibitem{xiao2025fairness}
Yisong Xiao, Aishan Liu, Siyuan Liang, Xianglong Liu, and Dacheng Tao.
\newblock Fairness mediator: Neutralize stereotype associations to mitigate bias in large language models.
\newblock In {\em ISSTA}, 2025.

\bibitem{xiao2025genderbias}
Yisong Xiao, Xianglong Liu, QianJia Cheng, Zhenfei Yin, Siyuan Liang, Jiapeng Li, Jing Shao, Aishan Liu, and Dacheng Tao.
\newblock Genderbias-vl: Benchmarking gender bias in vision language models via counterfactual probing: Y. xiao et al.
\newblock {\em International Journal of Computer Vision}, 2025.

\bibitem{park2024geometry}
Kiho Park, Yo~Joong Choe, Yibo Jiang, and Victor Veitch.
\newblock The geometry of categorical and hierarchical concepts in large language models.
\newblock In {\em The Thirteenth International Conference on Learning Representations}, 2025.

\bibitem{park2024linear}
Kiho Park, Yo~Joong Choe, and Victor Veitch.
\newblock The linear representation hypothesis and the geometry of large language models.
\newblock In {\em International Conference on Machine Learning}, pages 39643--39666. PMLR, 2024.

\bibitem{nanda2023emergent}
Neel Nanda, Andrew Lee, and Martin Wattenberg.
\newblock Emergent linear representations in world models of self-supervised sequence models.
\newblock In {\em Proceedings of the 6th BlackboxNLP Workshop: Analyzing and Interpreting Neural Networks for NLP}, pages 16--30, 2023.

\bibitem{liu2023towards}
Aishan Liu, Shiyu Tang, Xinyun Chen, Lei Huang, Haotong Qin, Xianglong Liu, and Dacheng Tao.
\newblock Towards defending multiple lp-norm bounded adversarial perturbations via gated batch normalization.
\newblock {\em International Journal of Computer Vision}, 2023.

\bibitem{liu2023exploring}
Aishan Liu, Shiyu Tang, Siyuan Liang, Ruihao Gong, Boxi Wu, Xianglong Liu, and Dacheng Tao.
\newblock Exploring the relationship between architecture and adversarially robust generalization.
\newblock In {\em CVPR}, 2023.

\bibitem{liu2021training}
Aishan Liu, Xianglong Liu, Hang Yu, Chongzhi Zhang, Qiang Liu, and Dacheng Tao.
\newblock Training robust deep neural networks via adversarial noise propagation.
\newblock {\em TIP}, 2021.

\bibitem{ouyang2022training}
Long Ouyang, Jeffrey Wu, Xu~Jiang, Diogo Almeida, Carroll Wainwright, Pamela Mishkin, Chong Zhang, Sandhini Agarwal, Katarina Slama, Alex Ray, et~al.
\newblock Training language models to follow instructions with human feedback.
\newblock {\em Advances in neural information processing systems}, 35:27730--27744, 2022.

\bibitem{dathathri2019plug}
Sumanth Dathathri, Andrea Madotto, Janice Lan, Jane Hung, Eric Frank, Piero Molino, Jason Yosinski, and Rosanne Liu.
\newblock Plug and play language models: A simple approach to controlled text generation.
\newblock In {\em International Conference on Learning Representations}, 2019.

\bibitem{liu2021dexperts}
Alisa Liu, Maarten Sap, Ximing Lu, Swabha Swayamdipta, Chandra Bhagavatula, Noah~A Smith, and Yejin Choi.
\newblock Dexperts: Decoding-time controlled text generation with experts and anti-experts.
\newblock In {\em Proceedings of the 59th Annual Meeting of the Association for Computational Linguistics and the 11th International Joint Conference on Natural Language Processing (Volume 1: Long Papers)}, pages 6691--6706, 2021.

\bibitem{krause2020gedi}
Ben Krause, Akhilesh~Deepak Gotmare, Bryan McCann, Nitish~Shirish Keskar, Shafiq Joty, Richard Socher, and Nazneen~Fatema Rajani.
\newblock Gedi: Generative discriminator guided sequence generation.
\newblock {\em arXiv preprint arXiv:2009.06367}, 2020.

\bibitem{xu2024genarm}
Yuancheng Xu, Udari~Madhushani Sehwag, Alec Koppel, Sicheng Zhu, Bang An, Furong Huang, and Sumitra Ganesh.
\newblock Genarm: Reward guided generation with autoregressive reward model for test-time alignment.
\newblock In {\em The Thirteenth International Conference on Learning Representations}, 2025.

\bibitem{wei2024assessing}
Boyi Wei, Kaixuan Huang, Yangsibo Huang, Tinghao Xie, Xiangyu Qi, Mengzhou Xia, Prateek Mittal, Mengdi Wang, and Peter Henderson.
\newblock Assessing the brittleness of safety alignment via pruning and low-rank modifications.
\newblock In {\em Proceedings of the 41st International Conference on Machine Learning}, pages 52588--52610, 2024.

\bibitem{wang2024detoxifying}
Mengru Wang, Ningyu Zhang, Ziwen Xu, Zekun Xi, Shumin Deng, Yunzhi Yao, Qishen Zhang, Linyi Yang, Jindong Wang, and Huajun Chen.
\newblock Detoxifying large language models via knowledge editing.
\newblock {\em arXiv preprint arXiv:2403.14472}, 2024.

\bibitem{gu2024model}
Jia-Chen Gu, Hao-Xiang Xu, Jun-Yu Ma, Pan Lu, Zhen-Hua Ling, Kai-Wei Chang, and Nanyun Peng.
\newblock Model editing harms general abilities of large language models: Regularization to the rescue.
\newblock In {\em Proceedings of the 2024 Conference on Empirical Methods in Natural Language Processing}, pages 16801--16819, 2024.

\bibitem{panickssery2023steering}
Nina Panickssery, Nick Gabrieli, Julian Schulz, Meg Tong, Evan Hubinger, and Alexander~Matt Turner.
\newblock Steering llama 2 via contrastive activation addition.
\newblock {\em arXiv preprint arXiv:2312.06681}, 2023.

\bibitem{liu2023context}
Sheng Liu, Haotian Ye, Lei Xing, and James Zou.
\newblock In-context vectors: Making in context learning more effective and controllable through latent space steering.
\newblock {\em arXiv preprint arXiv:2311.06668}, 2023.

\bibitem{ziegler2019fine}
Daniel~M Ziegler, Nisan Stiennon, Jeffrey Wu, Tom~B Brown, Alec Radford, Dario Amodei, Paul Christiano, and Geoffrey Irving.
\newblock Fine-tuning language models from human preferences.
\newblock {\em arXiv preprint arXiv:1909.08593}, 2019.

\bibitem{go2023aligning}
Dongyoung Go, Tomasz Korbak, Germ{\'a}n Kruszewski, Jos Rozen, Nahyeon Ryu, and Marc Dymetman.
\newblock Aligning language models with preferences through f-divergence minimization.
\newblock In {\em Proceedings of the 40th International Conference on Machine Learning}, pages 11546--11583, 2023.

\bibitem{meng2022locating}
Kevin Meng, David Bau, Alex Andonian, and Yonatan Belinkov.
\newblock Locating and editing factual associations in gpt.
\newblock {\em Advances in Neural Information Processing Systems}, 35:17359--17372, 2022.

\bibitem{shlens2014tutorial}
Jonathon Shlens.
\newblock A tutorial on principal component analysis.
\newblock {\em arXiv preprint arXiv:1404.1100}, 2014.

\bibitem{sutton1998reinforcement}
Richard~S Sutton, Andrew~G Barto, et~al.
\newblock {\em Reinforcement learning: An introduction}, volume~1.
\newblock MIT press Cambridge, 1998.

\bibitem{pignatellisurvey}
Eduardo Pignatelli, Johan Ferret, Matthieu Geist, Thomas Mesnard, Hado van Hasselt, and Laura Toni.
\newblock A survey of temporal credit assignment in deep reinforcement learning.
\newblock {\em Transactions on Machine Learning Research}.

\bibitem{merity2017pointer}
Stephen Merity, Caiming Xiong, James Bradbury, and Richard Socher.
\newblock Pointer sentinel mixture models.
\newblock In {\em International Conference on Learning Representations}, 2017.

\bibitem{Detoxify}
Laura Hanu and {Unitary team}.
\newblock Detoxify.
\newblock Github. https://github.com/unitaryai/detoxify, 2020.

\bibitem{gao2021framework}
Leo Gao, Jonathan Tow, Stella Biderman, Sid Black, Anthony DiPofi, Charles Foster, Laurence Golding, Jeffrey Hsu, Kyle McDonell, Niklas Muennighoff, et~al.
\newblock A framework for few-shot language model evaluation.
\newblock {\em Version v0. 0.1. Sept}, page~8, 2021.

\bibitem{clark2019boolq}
Christopher Clark, Kenton Lee, Ming-Wei Chang, Tom Kwiatkowski, Michael Collins, and Kristina Toutanova.
\newblock Boolq: Exploring the surprising difficulty of natural yes/no questions.
\newblock In {\em Proceedings of the 2019 Conference of the North American Chapter of the Association for Computational Linguistics: Human Language Technologies, Volume 1 (Long and Short Papers)}, pages 2924--2936, 2019.

\bibitem{wang2018glue}
Alex Wang, Amanpreet Singh, Julian Michael, Felix Hill, Omer Levy, and Samuel Bowman.
\newblock Glue: A multi-task benchmark and analysis platform for natural language understanding.
\newblock In {\em Proceedings of the 2018 EMNLP Workshop BlackboxNLP: Analyzing and Interpreting Neural Networks for NLP}, pages 353--355, 2018.

\bibitem{zellers2019hellaswag}
Rowan Zellers, Ari Holtzman, Yonatan Bisk, Ali Farhadi, and Yejin Choi.
\newblock Hellaswag: Can a machine really finish your sentence?
\newblock In {\em Proceedings of the 57th Annual Meeting of the Association for Computational Linguistics}, pages 4791--4800, 2019.

\bibitem{sakaguchi2021winogrande}
Keisuke Sakaguchi, Ronan~Le Bras, Chandra Bhagavatula, and Yejin Choi.
\newblock Winogrande: An adversarial winograd schema challenge at scale.
\newblock {\em Communications of the ACM}, 64(9):99--106, 2021.

\bibitem{clark2018think}
Peter Clark, Isaac Cowhey, Oren Etzioni, Tushar Khot, Ashish Sabharwal, Carissa Schoenick, and Oyvind Tafjord.
\newblock Think you have solved question answering? try arc, the ai2 reasoning challenge.
\newblock {\em arXiv preprint arXiv:1803.05457}, 2018.

\bibitem{mihaylov2018can}
Todor Mihaylov, Peter Clark, Tushar Khot, and Ashish Sabharwal.
\newblock Can a suit of armor conduct electricity? a new dataset for open book question answering.
\newblock In {\em Proceedings of the 2018 Conference on Empirical Methods in Natural Language Processing}, pages 2381--2391, 2018.

\bibitem{radford2019language}
Alec Radford, Jeffrey Wu, Rewon Child, David Luan, Dario Amodei, Ilya Sutskever, et~al.
\newblock Language models are unsupervised multitask learners.
\newblock {\em OpenAI blog}, 1(8):9, 2019.

\bibitem{zhang2022opt}
Susan Zhang, Stephen Roller, Naman Goyal, Mikel Artetxe, Moya Chen, Shuohui Chen, Christopher Dewan, Mona Diab, Xian Li, Xi~Victoria Lin, et~al.
\newblock Opt: Open pre-trained transformer language models.
\newblock {\em arXiv preprint arXiv:2205.01068}, 2022.

\bibitem{jiang2023mistral7b}
Albert~Q. Jiang, Alexandre Sablayrolles, Arthur Mensch, Chris Bamford, Devendra~Singh Chaplot, Diego de~las Casas, Florian Bressand, Gianna Lengyel, Guillaume Lample, Lucile Saulnier, Lélio~Renard Lavaud, Marie-Anne Lachaux, Pierre Stock, Teven~Le Scao, Thibaut Lavril, Thomas Wang, Timothée Lacroix, and William~El Sayed.
\newblock Mistral 7b, 2023.

\bibitem{tunstall2023zephyr}
Lewis Tunstall, Edward Beeching, Nathan Lambert, Nazneen Rajani, Kashif Rasul, Younes Belkada, Shengyi Huang, Leandro von Werra, Cl{\'e}mentine Fourrier, Nathan Habib, et~al.
\newblock Zephyr: Direct distillation of lm alignment.
\newblock {\em arXiv preprint arXiv:2310.16944}, 2023.

\bibitem{touvron2023llama}
Hugo Touvron, Louis Martin, Kevin Stone, Peter Albert, Amjad Almahairi, Yasmine Babaei, Nikolay Bashlykov, Soumya Batra, Prajjwal Bhargava, Shruti Bhosale, et~al.
\newblock Llama 2: Open foundation and fine-tuned chat models.
\newblock {\em arXiv preprint arXiv:2307.09288}, 2023.

\bibitem{khanovargs}
Maxim Khanov, Jirayu Burapacheep, and Yixuan Li.
\newblock Args: Alignment as reward-guided search.
\newblock In {\em The Twelfth International Conference on Learning Representations}.

\bibitem{wingate2022prompt}
David Wingate, Mohammad Shoeybi, and Taylor Sorensen.
\newblock Prompt compression and contrastive conditioning for controllability and toxicity reduction in language models.
\newblock In {\em Findings of the Association for Computational Linguistics: EMNLP 2022}, pages 5621--5634, 2022.

\bibitem{sun2024trustllm}
Lichao Sun, Yue Huang, Haoran Wang, Siyuan Wu, Qihui Zhang, Chujie Gao, Yixin Huang, Wenhan Lyu, Yixuan Zhang, Xiner Li, et~al.
\newblock Trustllm: Trustworthiness in large language models.
\newblock {\em arXiv preprint arXiv:2401.05561}, 3, 2024.

\bibitem{Phan2023}
Long Phan.
\newblock harmful harmless instructions.
\newblock HuggingFace. https://huggingface.co/datasets/justinphan3110/harmful\_harmless\_instructions, 2023.

\end{thebibliography}
}


\end{document}